\documentclass[runningheads]{llncs}
\usepackage{graphicx}
\usepackage{amsmath}

\usepackage[
separate-uncertainty = true,
multi-part-units = repeat
]{siunitx}
\usepackage[citebordercolor={1 1 1},linkbordercolor={1 1 1},urlbordercolor={1 1 1}]{hyperref}
\usepackage{floatrow}
\usepackage{flexisym}
\usepackage{multirow}
\usepackage{amsmath}
\usepackage{tabularx}
\usepackage{lineno}
\usepackage{diagbox}
\usepackage{wrapfig,lipsum,booktabs}
\usepackage{amssymb}

\usepackage[misc,geometry]{ifsym} 
\begin{document}
\title{SST-GNN: Simplified Spatio-temporal Traffic forecasting model using Graph Neural Network}

\titlerunning{SST-GNN: Simplified Spatio-temporal Traffic forecasting GNN}
%

\author{Amit Roy$^\star$\textsuperscript{(\Letter)}
\and
Kashob Kumar Roy\thanks{Equal Contribution}
\and\\
Amin Ahsan Ali
\and
M Ashraful Amin
\and 
A K M Mahbubur Rahman \\
\email{\{amitroy7781, kashobroy\}@gmail.com}, \\
\email{\{aminali, aminmdashraful, akmmrahman\}@iub.edu.bd}
}
\authorrunning{A. Roy et al.}
\institute{
Artificial Intelligence and Cybernetics Lab, Independent University, Bangladesh 
\\
\url{https://www.agencylab.org/} 
}

\maketitle              
\begin{abstract}
To capture spatial relationships and temporal dynamics in traffic data, spatio-temporal models for traffic forecasting have drawn significant attention in recent years. Most of the recent works employed graph neural networks(GNN) with multiple layers to capture the spatial dependency. However, road junctions with different hop-distance can carry distinct traffic information which should be exploited separately but existing multi-layer GNNs are incompetent to discriminate between their impact. Again, to capture the temporal interrelationship, recurrent neural networks are common in state-of-the-art approaches that often fail to capture long-range dependencies. Furthermore, traffic data shows repeated patterns in a daily or weekly period which should be addressed explicitly.  To address these limitations, we have designed a \textbf{S}implified \textbf{S}patio-temporal \textbf{T}raffic forecasting \textbf{GNN(SST-GNN)} that effectively encodes the spatial dependency by separately aggregating different neighborhood representations rather than with multiple layers and capture the temporal dependency with a simple yet effective weighted spatio-temporal aggregation mechanism. We capture the periodic traffic patterns by using a novel position encoding scheme with historical and current data in two different models. With extensive experimental analysis, we have shown that our model\footnote{Code is available at \href{https://github.com/AmitRoy7781/SST-GNN}{\color{magenta} github.com/AmitRoy7781/SST-GNN}} has significantly outperformed the state-of-the-art models on three real-world traffic datasets from the Performance Measurement System (PeMS).

\end{abstract}
\section{Introduction}
In recent years, future traffic prediction is getting interests among researchers from the area of Intelligent Transportation System(ITS). Generally, the traffic intensity of given sensors refers to the speed of people/vehicles passing
through those sensors on traffic networks at each timestamp. 
Accurate forecasting of future traffic speeds has plenty of advantages such as it would help citizens not only to bypass the crowded
path but also to schedule an efficient trip in advance. 
However, the task of traffic forecasting is challenging because the traffic in a busy metropolitan city changes across different locations throughout the different time periods every day. Also, different traffic patterns are observed on weekdays and weekends. Hence, there lies a complex spatio-temporal relationship in traffic data that makes the task of accurate traffic prediction challenging.

As the traffic network of a city can be modeled as a graph with traffic speed of different nodes (road junctions) across different timestamps, most of the recent approaches~\cite{li2017diffusion,wu2019graph,guo2019attention,park2019stgrat,chen2019gated,fang2019gstnet,ijcai2020-326,wu2019graph} have tried to design the problem of traffic forecasting as a regression task. In these models, the spatial relationship among different nodes are captured using graph neural networks (GNNs)~\cite{kipf2016semi,hamilton2017inductive} and recurrent neural networks are employed to consider the temporal dependency~\cite{yu2017spatio}. To mention a few, STGCN~\cite{yu2017spatio} is the first approach to apply graph convolution to capture spatial representation in traffic forecasting along with recurrent units for temporal dependencies. On the other hand, DCRNN~\cite{li2017diffusion} employed bi-directional random walk to preserve spatial relation and GRU for temporal dependencies.

In spite of the extensive efforts for future traffic prediction, the challenge is not solved yet due to a couple of reasons. Firstly, state-of-the-art models have a common practice to increase the receptive field by using multi-layer GNNs to capture the spatial traffic information from different-hop neighborhoods. However, the immediate neighboring junctions might have different impacts on the target node's traffic pattern from the distant junctions. Multi-layer GNNs
suffer from over-smooth problem~\cite{chen2020measuring} while aggregating the information from different hop neighboring junctions in more layers which results in
less informative spatial representations.  Instead,
 directly employing the representation of different-hop neighbors towards the fully connected layers will be more effective to encode the impact of different hop neighboring junctions~\cite{zhu2020beyond}. 
Secondly, traditional spatio-temporal models apply recurrent neural networks e.g., LSTM, GRU to encode the temporal information. However, recurrent neural networks often fail to perform well to forecast the traffic in long-range prediction as spatial traffic at different timestamps has a varying scale of impact on the target node's pattern.   To encode the temporal dependency explicitly,  we propose a novel spatio-temporal weighted aggregation scheme that can learn the importance of the spatial representation from previous timestamps. Also, we stack the representation of different timestamps to obtain the final representation that allows our model in handling long-range dependencies effectively. 

\begin{figure}[t]
\floatbox[{\capbeside\thisfloatsetup{capbesideposition={right,top},capbesidewidth=4cm}}]{figure}[\FBwidth]
{\caption{\small Predicting the traffic of 10:05 AM-11:00 AM on Tuesday by observing the traffic data of the past hour from the last seven days as well as the present day to capture the daily pattern and current day pattern.}
\label{fig:motivation_his_cur}}
{\includegraphics[width=6cm,height=3.5cm]{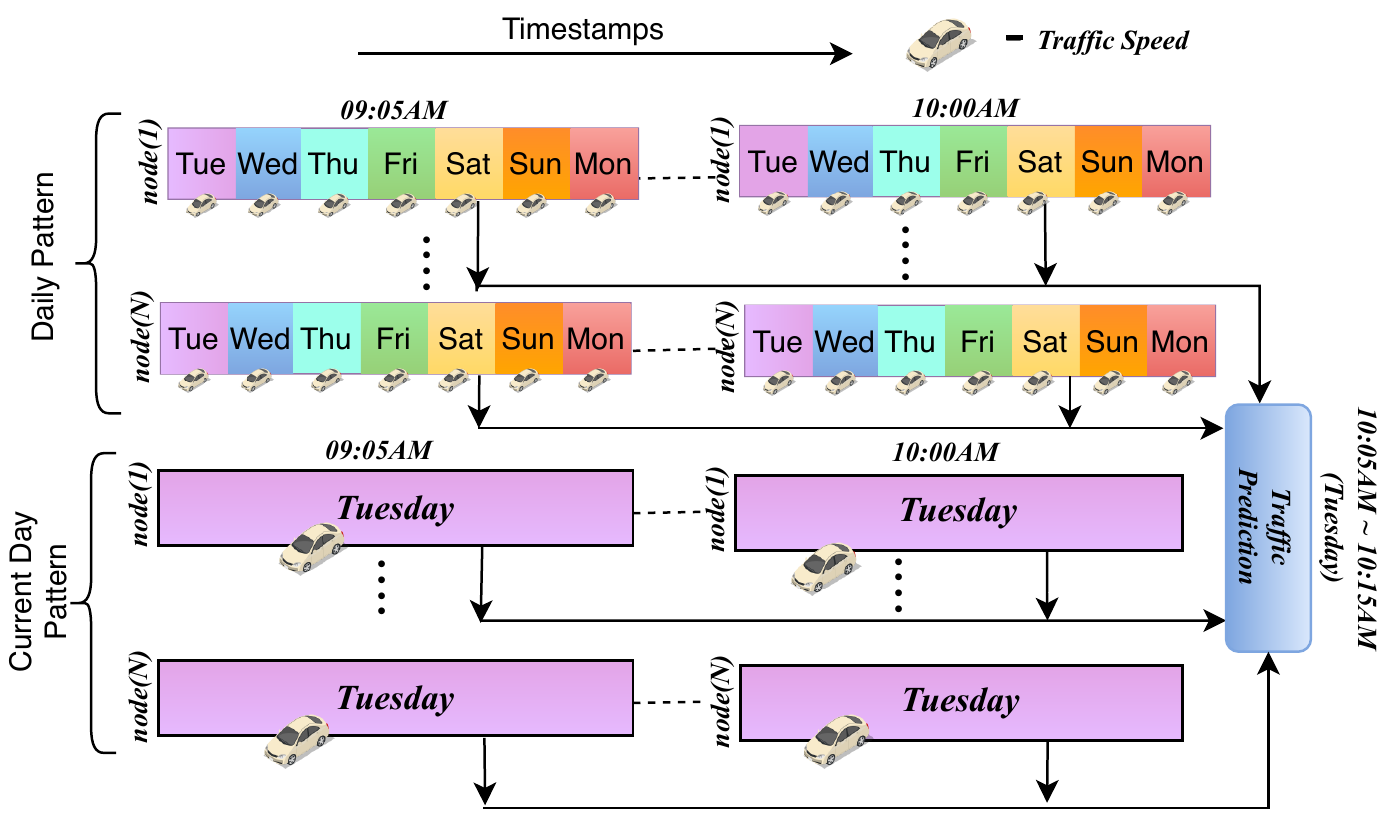}}
\end{figure}

Finally, traffic data shows repetitive daily patterns across days in a week. To learn these trends in traffic data, an ideal model should consider the current day pattern as well as the daily pattern seen in the traffic data.  Here, we define the current day pattern as the traffic situation  observed in the last hour on the current day and the daily pattern as the traffic intensities exist in the same time period in the last one week. Most of the researchers put their contribution to learning the current day pattern. For instance, to predict traffic speed at 10:05 AM - 11:00 AM on Tuesday, recent researchers propose frameworks to learn the pattern from 9:05 AM - 10:00 AM on the present day (Tuesday) which is depicted  as current day pattern in Figure \ref{fig:motivation_his_cur}. However, current day pattern information might not be enough to model city traffic.  In our work, we learn the traffic pattern effectively with two different models named as the current-day model and historical model where the current-day model analyze the past hour data on the current day and the historical model deals with the past hour traffic intensity in the last seven days (Figure \ref{fig:motivation_his_cur}). Lastly, the traffic intensity in a metropolitan city varies throughout different time periods in a day across weekdays and weekends. Therefore, we enhance the generalization capability of our model with a novel position encoding scheme which helps our model to distinguish between traffic data of different periods of the day on both weekdays and weekends.  In summary,  the key contribution of our work SST-GNN includes: 
\begin{itemize}

\item We directly utilize the representation of different hop neighbors rather than using multi-layer GNNs to explicitly focus on  the spatial dependency of traffic intensity from road junctions at different hop distance.

\item We capture the temporal dependency with a simple weighted aggregation of the spatial representations from the different timestamps and finally stacking them to capture inter-timestamp dependency.

\item  We propose a simple yet effective framework to extract current day and daily information through two different models: current-day model and historical model. The framework uses neighborhood aggregation based graph neural networks to learn the node embeddings.

\item We propose a position encoding scheme that can encode the periodic information of days and weeks into traffic data which can be easily extended to months and even for years.

\item From the extensive experimental analysis, we show the efficacy of our model. Our model SST-GNN outperforms the state-of-the-art models in predicting the traffic speed of the next 15, 30, 45, and 60 minutes.
\end{itemize}

\section{Background Study}
\textbf{Related works:} In the early years, various statistical and machine learning techniques such as Auto-Regressive Integrated Moving Average (ARIMA), Historical Average (HA), Support Vector Regression (SVR), and Kalman filters have been widely used for traffic forecasting. However, in recent years, graph neural networks(GNN) have achieved greater success in modeling real-life traffic. GNNs are able to encode the spatial dependency between neighbor nodes in a graph into their hidden representation by employing different feature aggregation scheme. Graph Convolution Networks~\cite{kipf2016semi,defferrard2016convolutional} apply spectral convolutions to learn structural dependency as well as feature information. On the other hand, GraphSAGE~\cite{hamilton2017inductive} introduced a neighborhood aggregation strategy to preserve the inter-relationship among proximal nodes. As GNNs succeeds in learning representations for various downstream machine learning tasks, several recent works have employed graph convolution to learn node representations that can extract spatial relations from the traffic network. STGCN\cite{yu2017spatio} has modeled spatial and temporal relations using a convolutional network.
The diffusion process is used to model the traffic networks in DCRNN\cite{li2017diffusion} that captures the spatial relations by using the bidirectional random walks and GRU for temporal dependencies. Besides, several recent works\cite{wu2019graph,fang2019gstnet,park2019stgrat} have achieved good performance.To capture the spatio-temporal dependency among nodes in the embedded space, Graph Wavenet~\cite{wu2019graph} learns a self-adaptive dependency matrix where the receptive field increases with the number of layers.  Very recent work LSGCN\cite{ijcai2020-326} proposes a new graph attention network
called cosAtt and incorporates the cosAtt and GCN into the spatial gated block and linear gated block to iteratively predict future traffic intensity.  We observe that state-of-the-art models fail to capture the impact of different hop neighborhoods for a targer node in traffic networks explicitly. Also, the RNN-based models are incompetent to learn temporal dependencies in long term prediction. To address the above challenges, we explicitly capture the impact of different-hop neighborhoods on target node's traffic with a simple yet effective spatio-temporal aggregation scheme and stack the embeddings of intermediate timestamps to learn temporal dependencies across different timestamps. Capturing the traffic of different hop neighborhood with simplified spatio-temporal aggregation improves our models performance than the state-of-the-art traffic forecasting models.

\noindent
\textbf{Preliminaries and Problem Definition:}
A traffic network is represented as a graph \textit{G = (V,A)} where \textit{V} is the set of nodes that denote road junctions and \textit{A} $\in$  $\mathbb{R}^{|V| \times|V|}$ is the adjacency matrix of the graph, where $A_{i,j}$ = 1 if junction $i$ and $j$ are connected by an road and 0 otherwise. Each node also contains some features of a junction representing traffic flow, speed, occupancy etc. As traffic at different nodes change over time, the traffic features  of a node \textit{u} at timestamp \textit{t} is denoted as $X_{u}^{<t>}$ $\in \mathbb{R}^{d}$ where $d$ denotes the feature dimension and $X^{<t>}$ $\in$ $\mathbb{R}^{|V| \times d}$ represents the traffic features of all nodes at timestamp \textit{t}. The graph at a timestamp $t$ is denoted as timestamp graph $G^{<t>}$.Note that, all timestamp graphs are structurally identical to each other. However, a traffic forecasting framework takes a sequence of $T$ timestamp graphs with their node features $(X^{<1>},X^{<2>},\ldots \ldots, X^{<T>})$ as input and predicts the traffic intensities of nodes at next $n$ timestamps that is (${Y}^{<T+1>}, {Y}^{<T+2>}, \ldots \ldots, {Y}^{<T+n>}$).

\begin{figure*}[!t]
\centering
\scalebox{0.85}{
\includegraphics[width=1.0\columnwidth]{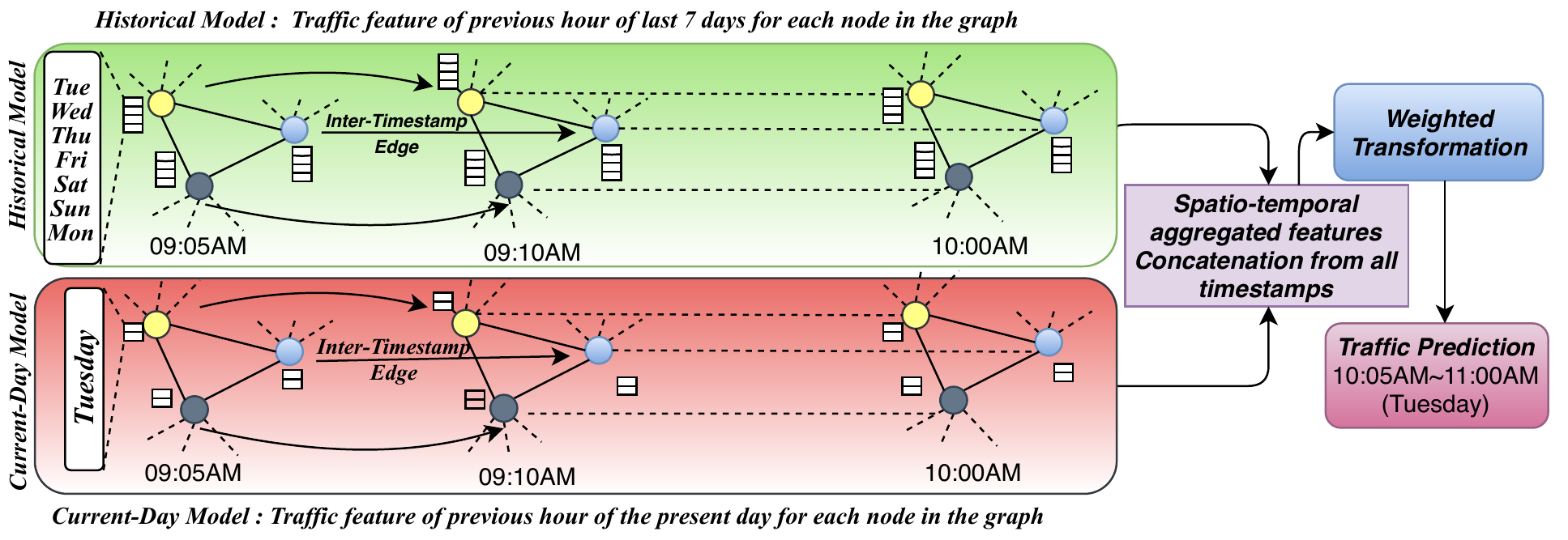}}
\caption{\small
Inter-timestamp edges are introduced between identical nodes of consecutive timestamps e.g. an edge between a blue node at timestamp 09:05 AM and a blue node at timestamp 09:10 AM where the same color indicates identical nodes.
Although both historical and current-day model deals with the same spatio-temporal graph consisting of all timestamp graphs over 5 min interval in the past hour of the prediction window, the historical model considers traffic features from last week to capture the repeated daily patterns while the current-day model uses only current day (e.g. Tuesday) information to find current day patterns in traffic data. Spatial dependency is captured through aggregating features from different neighborhoods on each timestamp graph while temporal dependency is preserved by performing temporal aggregation among the node representations learned from previous timestamps which are depicted in Fig.~\ref{fig:traffic_model_weighted_aggregation}. Finally, concatenation followed by weighted transformation is performed to compute the spatio-temporal embeddings of nodes which are used for traffic prediction.
}
\label{fig:traffic_model_overview}
\end{figure*}

\section{Proposed Model}

In this section, we describe the whole architecture of our proposed framework that can effectively capture spatio-temporal dependencies between road junctions. We discuss spatio-temporal graph and positional encoding scheme for performing spatio-temporal aggregation and capturing periodicity in traffic data respectively. After that, we present spatio-temporal aggregation with two different models namely historical model and current-day model and concluded with the final embedding and training process.  A high-level overview has been presented in Fig.~\ref{fig:traffic_model_overview} and Fig.~\ref{fig:traffic_model_weighted_aggregation}.

\noindent
\textbf{Spatio-Temporal Graph:} To capture the complex spatio-temporal dependencies between nodes across different timestamp graphs, we introduce inter-timestamp edges between identical nodes of consecutive timestamp graphs as shown in Fig.~\ref{fig:traffic_model_overview} where the same color indicates identical nodes. Afterward, to learn embeddings of nodes, we perform our proposed spatio-temporal aggregation on a spatio-temporal graph that consists of previous $T$ timestamp graphs from the prediction window with their inter-timestamp edges.  

\noindent
\textbf{Positional Encoding:}
To extract informative traffic features from different periods of the day, we need to encode the relative position of the different time periods in our model. Following the relative positioning concept widely used in transformer based attention mechanism in Machine Translation~\cite{vaswani2017attention}, we have used positional encoding with a sinusoidal function to provide position information on different timestamps. We ensure that the sinusoidal function for each day completes a full cycle within a day. Hence, any time duration can be represented as a repetitive portion of the sine curve of each day.  For example, the sinusoidal curve will have the same pattern during the time slot (9:05 AM - 10:00 AM) daily. Hence, this positional encoding will help the model capture daily pattern indeed. Moreover, there might be a weekly pattern in traffic such as specific days that might have the same kind of traffic. Also, the proposed framework needs to see whether the patterns are coming from weekdays or weekends. To capture this kind of weekly pattern, we also propose another full cycle of a sine wave for each week. Therefore, the final position encoding has been achieved by Eq.~\ref{eq:position_encoding}.
\begin{equation}
    \mathcal{P}^{<t>} = sin(\frac{ 2\pi t}{24 \times hr\_sample}) + sin(\frac{2 \pi t}{24\times 7 \times hr\_sample})
    \label{eq:position_encoding}
\end{equation}
where $t$ denotes a particular timestamp and $hr\_sample$ represents the number of observed data samples in an hour. The idea can be extended to capture monthly repetition with another full cycle sine wave that completes in a month. 

\noindent
\textbf{Spatio-Temporal Aggregation:}
We develop a spatio-temporal aggregation scheme to encode spatial as well as temporal dependencies into the embeddings of nodes that have been shown in Fig~\ref{fig:traffic_model_weighted_aggregation}. It has two components as follows: 
\begin{itemize}
   
    \item \textbf{Spatial Aggregation}: In real-life traffic networks, it can be observed that all higher-order neighborhoods are not equally important for a target node. Different hop neighborhood may carry distinct information that should be captured explicitly. Therefore, we perform information aggregation over nodes in different neighborhoods separately in each timestamp graph as follows,
 
    \begin{equation}
    X_{(k)}^{<t>} = D_{(k)}^{-1} A_{(k)}X^{<t>}; \hspace*{2em} S_{u}^{<t>} = \sum_{k=1}^K  X_{(k),u}^{<t>} W_{(k)}^{<t>}
    \label{eq:spatial_aggregation}
    \end{equation}
    where, $A_{(k)}$ denotes $k^{th}$-hop neighborhood - meaning that $|A_{(k)}|_{i,j} = 1$ only if node $i$ and $j$ are exactly $k$ hop away from each other otherwise 0, $D_{(k)}$ is the degree matrix of $A_{(k)}$, $X_{(k)}^{<t>}$ is the degree-normalized mean of $k^{th}$-hop neighbor-embeddings at timestamp $t$, Further, we perform weighted aggregation among the mean representations of different-hop  neighborhoods up to $K$ hop away from node $u$ to compute the spatial embeddings of node $u$ denoted as $S_{u}^{<t>}$ where $W_k^{<t>}$ is learnable weight parameters to capture the impact of $k^{th}$-hop neighborhood at timestamp $t$. Explicit aggregation of different hop neighborhood embeddings helps to differentiate the impacts of the traffic intensities from different hop neighbor nodes on a target node.

\begin{figure*}[b]
\centering
\scalebox{0.85}{
\includegraphics[width=1.0\columnwidth]{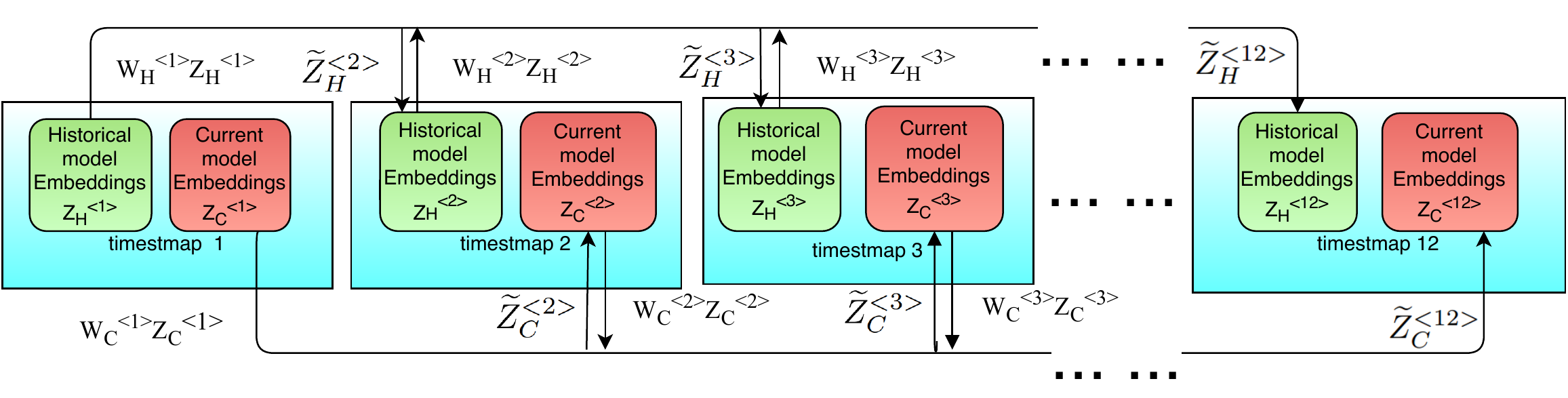}}
\caption{Spatio-Temporal Aggregation Scheme: To capture complex spatio-temporal dependencies in traffic networks, the historical model concatenates the spatial embeddings from different hop neighborhoods at timestamp $t$ with temporal embedding $\tilde{Z}_{H}^{t}$ - the weighted aggregation of (${Z}_{H}^{1},\dots,{Z}_{H}^{t-1}$), to learn spatio-temporal embeddings ${Z}_{H}^{t}$. Similarly, current-day model performs the same process.}
\label{fig:traffic_model_weighted_aggregation}
\end{figure*}    
    
\item \textbf{Temporal Aggregation}: To capture temporal dynamics among different timestamp graphs, temporal embeddings of nodes, $\tilde{Z}_{u}^{<t>}$, at timestamp $t$ are computed through aggregating spatio-temporal embeddings from the earlier timestamps as follows,
\begin{equation}
\tilde{Z}_{u}^{<t>} = ReLU(\sum_{i = 1}^{t-1}(W^{<i>}Z_{u}^{<i>}))
\label{eq:upto_timestamp_embedding}
\end{equation}
 where $Z_u^{<i>}$ is the spatio-temporal embedding of $u$
and $W^{<i>}$ is the learnable weight at timestamp $i$.

\end{itemize}

After that we concatenate the ego(target node), spatial and temporal embeddings of node $u$ to learn the spatio-temporal embedding of node $u$ at timestamp $t$, $Z_{u}^{<t>}$ as following,
\begin{equation}
    Z_{u}^{<t>} = ReLU(W^{<t>}_{\tiny sptemp} ( X_u^{<t>} \parallel S_{u}^{<t>}  \parallel \tilde{Z}_{u}^{<t>})) + \mathcal{P}^{<t>}
    \label{eq:spatio_temporal_aggregation}
\end{equation}
 where $W_{sptemp}$ is a learnable parameter at timestamp $t$ and $\parallel$ denotes concatenation operation while $\mathcal{P}^{<t>}$ represents the positional encoding of timestamp $t$. In Equation~\ref{eq:spatio_temporal_aggregation}, 
temporal embedding of $u$ at timestamp $t$, $ \tilde{Z}_{u}^{<t>}$ captures the temporal dependencies of traffic from previous $1$ to $t$ - $1$ timestamps while spatial embedding $S_{u}^{<t>}$ leverages  information from different hop neighborhoods of node $u$. Moreover, our model can achieve its best generalization ability by keeping the ego(target node), spatial and temporal information separate without mixing them.  Furthermore, the periodic information of traffic data is also preserved by incorporating the positional encoding value of timestamp $t$ into node embeddings. Therefore, Equation~\ref{eq:spatio_temporal_aggregation} ensures that our model can learn complex traffic flow information across different hop neighbor road junctions as well as from different timestamps effectively.

\noindent
\textbf{Historical Model:} 
To preserve the historical traffic information of previous days, we propose a novel historical model that analyzes the daily patterns. In the historical model, we assign the feature vector of node $u$, $X^{<t>}_{H_{u}}$ $\in$ $\mathbb{R}^{P}$ as the traffic speed  at timestamp $t$ of last $P$ days. Therefore, the historical model captures the traffic pattern of the last $P$ days of previous $T$ timestamps from the prediction window. The motivation behind using the historical model is to capture the periodic nature of traffic data from the history of the last $P=7$ days. On each timestamp $t$, we perform spatio-temporal aggregation to learn historical spatio-temporal embedding  $Z_{H_u}^{<t>}$  for  each node $u$ as shown in Fig.~\ref{fig:traffic_model_weighted_aggregation}. 

\noindent
\textbf{Current-Day Model:} 
The current-day model only considers the traffic speed at timestamp $t$ of current day, $X^{<t>}_{C_{u}}$ $\in$ $\mathbb{R}$ as the feature vector of each node in the network just like the traditional traffic forecasting frameworks. Hence, the current-day model focuses on the last $T$ timestamps of the present day (prediction day) to capture the traffic pattern on the current day.  Similar to the historical model, in our current-day model we also perform spatio-temporal aggregation on each timestamp network to find current day spatio-temporal embedding $Z_{C_u}^{<t>}$  for node $u$  at timestamp $t$ that has been shown in Fig~\ref{fig:traffic_model_weighted_aggregation}.

\noindent
\textbf{Final Embedding:}
 After obtaining the desired embeddings for node $u$ by applying spatio-temporal aggregation for $ T = 12$ timestamps in the historical and current-day model, the embeddings from both models are concatenated and combined into final embedding  $Z_{F_u}$ for each node $u$ in input traffic network as follows,
\begin{equation}
    \widetilde{Z}_{F_{u}} = Z_{H_u}^{<1>}\parallel \ldots \parallel Z_{H_u}^{<T>} \parallel Z_{C_u}^{<1>} \parallel  \ldots \parallel Z_{C_u}^{<T>} 
    \label{eq:all_timestamps_embeddings}
    \end{equation}
    \begin{equation}
     {Z}_{F_{u}} = W_{F} . \widetilde{Z}_{F_{u}}
\end{equation}

where $Z_{H_u}^{<t>}$ and $Z_{C_u}^{<t>}$ represents the spatio-temporal embeddings from historical and current-day models respectively  for node $u$  at timestamp $t$ and $W_{F}$ is the learnable weight parameter. Combining the embeddings from all timestamps in Eq.~\ref{eq:all_timestamps_embeddings} enables our model to gain more expressiveness, in contrast existing models only focus on the embedding from last timestamp that limits the expressiveness to some extent. Finally, we have used a two-layer neural network to predict the traffic intensities at different nodes and update all the parameters by optimizing supervised mean squared error(MSE) as the loss function. 

\section{Experimental Analysis}
In this section, we describe datasets, dataset preprocessing, and experiment setup followed by the elaborate analysis of observed results.

\noindent
\textbf{Dataset Description:}
To prove the effectiveness of our proposed model, we have conducted experiments on three publicly available real-life traffic datasets  PeMSD7, PeMSD4, and PeMSD8~\cite{ijcai2020-326}
that are widely used for performance comparison in previous works such as STGCN~\cite{yu2017spatio}, ASTGCN\cite{guo2019attention}, LSGCN\cite{ijcai2020-326}. PeMSD7 contains the traffic data of California that consists of the traffic speed of 228 sensors with 832 road segments while the time span is from May, 2012 to June, 2012 (only weekdays). We choose the first month of traffic data as the training set while the rest are split equally into validation and test set. PeMSD4 consists of the traffic data of San Francisco with 307 sensors on 340 roads. The time span of the dataset is January-February in 2018 and we choose the first 47 days as the training set while the rest are used as validation and test set. Lastly,  PeMSD8 consists of the traffic data from San Bernardino with 170 detectors on 295 roads, ranging from July to August in 2016. We select the first fifty days as the training and the rest are used as the validation and test set. All three datasets contain traffic feature with an interval of five minutes. In all the experiments, we consider traffic speed as the traffic feature for all three datasets.

\noindent
\textbf{Data Preprocessing:}
Adjacency matrix of the sensor  network is constructed using a thresholded Gaussian kernel, 
$A_{ij}$=1 only if $i\neq j$ and $exp(- \frac{d^{2}_{ij}}{\delta}) \geq \epsilon$, otherwise 0 where $A_{ij}$ 
determines edge between sensor $i$ and $j$ which is related with $d_{ij}$ (the distance between sensor $i$ and $j$). To control the distribution and sparsity of adjacency matrix $A$, we set the thresholds $\delta = 0.1$ and $\epsilon = 0.5$

\noindent
\textbf{Experimental Settings:}
The experiments are conducted on a Linux computer (GeForce RTX2080 Ti GPU) where both historical and current-day model adopts 60 minutes time window i.e previous 12 timestamps are used to predict  traffic of the next 15, 30, 45, and 60 minutes. In historical model, the input feature vector of each node comprises the traffic speed of the last seven days while the current-day model considers the traffic speed of the current day in the corresponding timestamp. For PeMSD7, we aggregate spatial information from the 2-hop neighborhood while 4-hop neighbors are considered for the other two datasets.
We train our model by minimizing Mean Square Error (MSE) as the loss function with ADAM optimizer for 500 epochs. For all the datasets, we set the initial learning rate 0.001 with a decay rate of 0.5 every seven epochs. To report the performance comparison among different models, we opt Mean Absolute Errors (MAE), Root Mean Squared Errors (RMSE) and Mean Absolute Percentage Errors (MAPE) as the evaluation metrics.

\subsection{Experiment Results}

\begin{table*}[!b]
\centering
\scalebox{0.70}{
\begin{tabular}{c|c|ccc|ccc|ccc|ccc}
\hline
\multirow{2}{*}{Datasets} & \multirow{2}{*}{Models} & \multicolumn{3}{c|}{\textit{15 min}}          & \multicolumn{3}{c|}{\textit{30 min}}          & \multicolumn{3}{c|}{\textit{45 min}}          & \multicolumn{3}{c}{\textit{60 min}}          \\ \cline{3-14} 
                          &                         & MAE           & RMSE          & MAPE          & MAE           & RMSE          & MAPE          & MAE           & RMSE          & MAPE          & MAE           & RMSE          & MAPE          \\ \hline

\multirow{6}{*}{PeMSD7}  

                          & DCRNN (2018)                  & 2.22          & 4.25          & 5.16          & 3.04          & 6.02          & 7.46          & 3.64          & 7.24          & 9.00          & 4.15          & 8.20          & 10.82         \\  
                          & STGCN  (2018)                 & 2.24          & 4.01          & 5.28          & 3.04          & 5.74          & 7.46          & 3.61          & 6.85          & 9.26          & 4.08          & 7.69          & 10.23         \\ 
                          & ASTGCN  (2019)                & 2.85          & 5.15          & 7.25          & 3.35          & 6.12          & 8.67          & 3.70          & 6.77          & 9.73          & 3.96          & 7.20          & 10.53         \\  
                        & Graph WaveNet (2019)                  & \underline{2.17}          & \underline{3.87}          & \underline{4.85}          & \underline{2.90}          & \underline{5.40}          & \underline{6.86}          & \underline{3.23}          & \underline{6.29}          & \underline{8.06}          & \underline{3.75}       & \underline{7.02}          & \underline{9.58} \\
                          & LSGCN (2020)                   & {2.22}          & {3.98}          & {5.14}          & {2.96}          & {5.47}          & {7.18}          & {3.43}          & 
                          {6.39}          & 
                          {8.51}          & 
                          {3.81}          & 
                          {7.09}          & 
                          {9.60}          \\

                          & \textbf{SST-GNN}(ours) & \textbf{2.04} & \textbf{3.53} & \textbf{4.77} & \textbf{2.67} &
                    \textbf{4.80} & \textbf{6.66} &
                    \textbf{3.17} &
                    \textbf{5.79} &
                    \textbf{8.00} &
                    \textbf{3.48} & \textbf{6.39} &
                    \textbf{9.04} 
\\ \hline
\multirow{6}{*}{PeMSD4}   

                          & DCRNN (2018) & 1.35 & 2.94 & 2.68 & 1.77 & 4.06 & 3.71 & 2.04 & 4.77 & 4.78 & 2.26 & 5.28 & 5.10 \\ 
                          & STGCN (2018)  & 1.47 & 3.01 & 2.92 & 1.93 & 4.21 & 3.98 & 2.26 & 5.01 & 4.73 & 2.55 & 5.65 & 5.39\\  
                          & ASTGCN (2019) & 2.12 & 3.96 & 4.16 & 2.42 & 4.59 & 4.80 & 2.60 & 4.97 & 5.20 & 2.73 & 5.21 &5.46 \\ 
                          & Graph WaveNet (2019) & \underline{1.30} & \underline{2.68} &\underline{2.67} & \underline{1.70} & \underline{3.82} & \underline{3.73} &  \underline{1.95} & \underline{4.16} & \underline{4.25} & \underline{2.03} & \underline{4.65} & \underline{4.60}\\
                          & LSGCN (2020)  &{1.45} & {2.93} & {2.90} & {1.82} & {3.92} & {3.84} & {2.04} & {4.47} & {4.42} & {2.22} & {4.83} & {4.85} \\ 
                           & \textbf{SST-GNN}(ours) & \textbf{1.23} & \textbf{2.53} & \textbf{2.37} & \textbf{1.82} & \textbf{3.47} & \textbf{3.69} & \textbf{1.84} & \textbf{3.86} & \textbf{3.93} & \textbf{2.13} & \textbf{4.45} & \textbf{4.69}  \\ \hline
\multirow{6}{*}{PeMSD8}  
 
                          & DCRNN (2018)                  & {1.17}          & 2.59          & 2.32          & 1.49          & 3.56          & 3.21          & 1.71          & 4.13          & 3.83          & 1.87          & 4.50          & 4.28          \\ 
                          & STGCN (2018)                  & 1.19          & 2.62          & 2.34          & 1.59          & 3.61          & 3.24          & 1.92          & 4.21          & 3.91          & 2.25          & 4.68          & 4.54          \\  
                          & ASTGCN (2019)                & 1.49          & 3.18          & 3.16          & 1.67          & 3.69          & 3.59          & 1.81          & 3.92          & 3.98          & 1.89          & 4.13          & 4.22          \\ 
                          & LSGCN (2020)                  & \underline{1.16}          & \underline{2.45}          & \underline{2.24}          & \underline{1.46}          & 
                          \underline{3.28}          & 
                          \underline{3.02}          & 
                          \underline{1.66}          & \underline{3.75}          & \underline{3.51}          & 
                          \underline{1.81}          & 
                          \underline{4.11}          & 
                          \underline{3.89}          \\

                           & \textbf{SST-GNN}(ours) &
                           \textbf{1.03} &
                           \textbf{2.08} &
                           \textbf{1.86} &
                           \textbf{1.39} &
                    \textbf{2.80} & \textbf{2.67} &
                    \textbf{1.62} &
                    \textbf{3.28} &
                    \textbf{3.20} &
                    \textbf{1.74} &
                    \textbf{3.57} &
                    \textbf{3.50} \\ \hline
\end{tabular}
}
\caption{Performance comparison in traffic prediction (\textbf{Best}, \underline{2nd Best})}
\label{tab:performance_comparison}
\end{table*}

\subsubsection{Comparison with baselines:}
In Table~\ref{tab:performance_comparison}, we present the performance comparison of our model named SST-GNN with the state-of-the-art models STGCN, DCRNN, ASTGCN, Graph WaveNet and LSGCN in 15, 30, 45, and 60 minutes traffic prediction. 
In Table~\ref{tab:performance_comparison}, it is easy to observe that our model outperforms all baseline models in both long and short-term predictions for all three evaluation metrics on PeMSD7, PeMSD4, and PeMSD8. The second-best performance has been observed for the recent work
Graph Wavenet in dataset PeMSD7, PeMSD4, and for LSGCN in PeMSD8.
Graph Wavenet learns an adaptive adjacency matrix with different granularity whereas LSGCN analyzes long-term and short-term patterns explicitly by employing attention-guided GCN and GLU. It is obvious that our model is able to capture complex spatio-temporal relationship more accurately through the proposed spatio-temporal aggregation scheme to outperform both the Graph Wavenet and LSGCN with reasonable margins. A number of architectural factors facilitate these improvements. Firstly, keeping the representations from different neighboring junctions separate allows the proposed model to learn the impact of different hop neighbors on the target node's traffic. Moreover,  our model captures the important historical pattern (daily pattern) by analyzing the data from the last seven days. The historical module helps our proposed framework in both long-term and short-term prediction with significantly better performance than Graph Wavenet, LSGCN, and other models.   Thirdly, weighted/attention based aggregation of the representations from the different time stamps facilitates long-term prediction. Careful observations of Table \ref{tab:performance_comparison} reveals that our model achieves significant performance in long-term predictions (45, and 60 minutes) for all three datasets. Finally, the position encoding helps our model to distinguish between different patterns that existed in different parts of the day.

\begin{table}[t]
\centering
\scalebox{0.75}{
\begin{tabular}{c|ccc|ccc|ccc|ccc}
\hline
\multirow{2}{*}{Models}                                                   & \multicolumn{3}{c|}{\textit{15 minutes}}                                                                     & \multicolumn{3}{c|}{\textit{30 minutes}}                                                                     & \multicolumn{3}{c|}{\textit{45 minutes}}                                                                     & \multicolumn{3}{c}{\textit{60 minutes}}                                                                     \\ \cline{2-13} 
                                                                                   & MAE                           &
                                                                                   RMSE                           & MAPE                               & MAE                                & RMSE                               & MAPE                               & MAE                                & RMSE                               & MAPE                               & MAE                                & RMSE                               & MAPE                               \\ \hline
                        
Current-Day only                                                         & 1.22                               & 2.62                               & 2.35                               & 1.44                               & 2.87                               & 2.76                               & 1.98                               & 3.77                               & 3.87                               & 2.29                               & 4.06                               & 4.51                               \\ \hline 
Historical only                                                          & 1.93                               & 3.94                               & 3.85                               & 2.21                               & 4.23                               & 4.36                               & 2.24                               & 4.26                               & 4.41                               & 2.47                               & 4.55                               & 4.70                               \\ \hline 
\begin{tabular}[c]{@{}c@{}}\textbf{SST-GNN}\end{tabular} & \multicolumn{1}{l}{\textbf{1.03}} & \multicolumn{1}{l}{\textbf{2.08}} & \multicolumn{1}{l|}{\textbf{1.86}} & \multicolumn{1}{l}{\textbf{1.39}} & \multicolumn{1}{l}{\textbf{2.80}} & \multicolumn{1}{l|}{\textbf{2.67}} & \multicolumn{1}{l}{\textbf{1.62}} & \multicolumn{1}{l}{\textbf{3.28}} & \multicolumn{1}{l|}{\textbf{3.20}} & \multicolumn{1}{l}{\textbf{1.74}} & \multicolumn{1}{l}{\textbf{3.57}} & \multicolumn{1}{l}{\textbf{3.50}} \\ \hline
\end{tabular}
}
\caption{Performance Comparison of Historical Model, Current-Day Model with {SST-GNN}(combined model) on PeMSD8 }
\label{tab:ablation_studies}
\end{table}

\begin{table*}[b]
\centering
\scalebox{0.70}{
\begin{tabular}{c|ccc|ccc|ccc|ccc}
\hline
\multirow{2}{*}{Models} & \multicolumn{3}{c|}{\textit{15 min}} & \multicolumn{3}{c|}{\textit{30 min}} & \multicolumn{3}{c|}{\textit{45 min}} & \multicolumn{3}{c}{\textit{60 min}} \\ \cline{2-13} 
 & MAE & RMSE & MAPE & MAE & RMSE & MAPE & MAE & RMSE & MAPE & MAE & RMSE & MAPE \\ \hline
\begin{tabular}[c]{@{}c@{}} LSGCN \\ (trained with {PeMSD8}; \\tested on {PeMSD8}) \end{tabular} & 1.16 & 2.45 & 2.24 & \textbf{1.46} & 3.28 & 3.02 & 1.66 & 3.75 & 3.51 & \textbf{1.81} & 4.11 & 3.89 \\ \hline
\begin{tabular}[c]{@{}c@{}} \textbf{SST-GNN}(ours)  \\(trained with  \textbf{PeMSD7};\\ tested on \textbf{PeMSD8})\end{tabular} & \textbf{1.14} & \textbf{2.12} & \textbf{2.07} & \textbf{1.46} & \textbf{2.76} & \textbf{2.71} & \textbf{1.54} & \textbf{3.15} & \textbf{3.00} & 1.94 & \textbf{3.69} & \textbf{3.74} \\ \hline
\end{tabular}
}
\caption{SST-GNN's performance on PeMSD8 while trained on PeMSD7}
\label{tab:trained_PeMSD7_on_PeMSD8_vs_LSGCN}
\end{table*}

\noindent
\textbf{Ablation study on contributions  from current-day and historical models:}  We perform ablation analysis to determine which part of the model brings the main performance gain. In Table~\ref{tab:ablation_studies}, we present the performance comparison among current-day model, the historical model, and  the combined SST-GNN model on PeMSD8. We observe that the performance of current-day model is competitive with the state-of-the-art models showing the effectiveness of spatio-temporal aggregation scheme with current-day traffic data. Though only the current-day model or historical model can not outperform the baselines, the combined model achieves significant performance gain,  demonstrating the significance of both historical and current day traffic data on performance gain.

\noindent
\textbf{Generalization Ability:}
To observe the generalization ability of SST-GNN, we train it with PeMSD7 and test it on PeMSD8. Particularly, the  PeMSD7 dataset doesn't include any weekends. However, the PeMSD8 dataset is comparatively large and it contains both weekdays and weekends. We compare our performance with LSGCN where the LSGCN has been solely trained and tested with PeMSD8. From Table \ref{tab:trained_PeMSD7_on_PeMSD8_vs_LSGCN}, it is easy to notice that our proposed model's performance (trained with  PeMSD7; tested on PeMSD8)  outperforms  LSGCN while even the LSGCN is trained and tested with the PeMSD8. The only exceptions are the MAEs for 30 minutes (equal MAEs) and 60 minutes. The results demonstrate that though the PeMSD7 does not have weekends, the SST-GNN with positional encoding allows proper attentional weights towards historical weekdays and weekends as well as the current day pattern for the test data.
\begin{figure*}[!t]
\centering
\scalebox{0.85}{
\includegraphics[width=1.0\columnwidth]{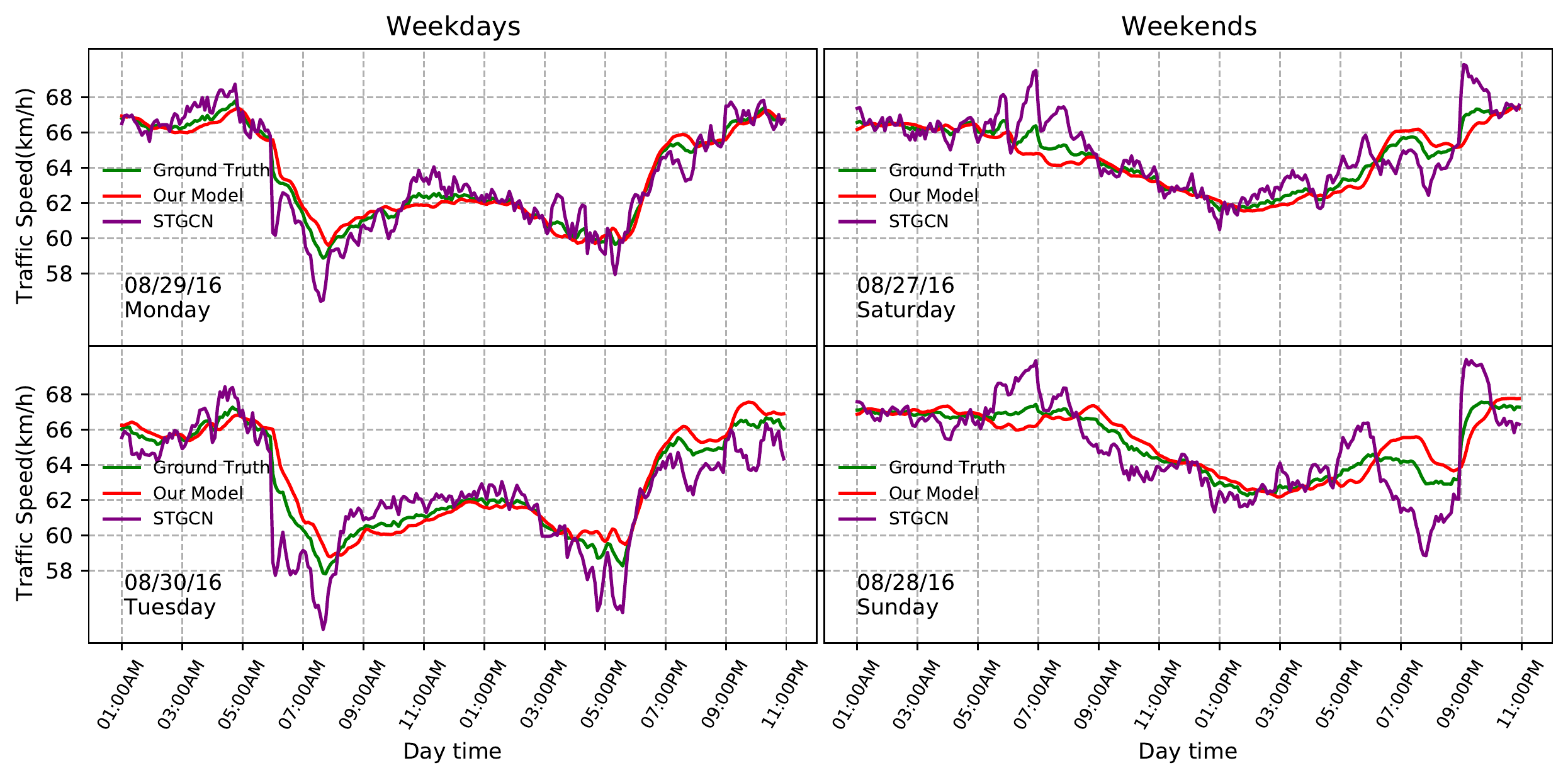}
}
\caption{Different periodic daily patterns on weekdays and weekends on PeMSD8. On the left, we can see speed decreases in morning peak and evening rush hours on weekdays whereas different traffic patterns are present on weekends.}
\label{fig:two_day_pattern}
\end{figure*}

\noindent
\textbf{Traffic Periodicity  on weekends and weekdays:}
In Figure~\ref{fig:two_day_pattern}, we plot the traffic speed of two consecutive weekdays and weekends from the PeMSD8 dataset to show how our model has learned the daily periodicity. The left column of plots demonstrates ground truths and predictions for two consecutive weekdays whereas the right column of plots depicts the ones for two consecutive weekends. In Figure~\ref{fig:two_day_pattern}, we can notice that our model can capture the daily periodicity and generalize among different time periods of the weekdays and weekends performing better than STGCN as our models prediction curve is more close to ground truth. In other words, the model can sufficiently distinguish the daily patterns between weekdays and weekends while capturing the historical and current-day patterns. Particularly,  the model captures the normal weekend patterns with slower traffic around the afternoon (previous weekend). It can also generalize sufficiently well in morning peaks and evening rush hours for weekdays as it can see the periodicity information from past weekdays through positional encoding.
\section{Conclusion}
Traffic data include repeated patterns on a daily and weekly basis. To capture the periodicity in traffic data, we design a novel spatial-temporal traffic forecasting framework that includes two different models namely historical and current-day model. The historical patterns are captured by observing the traffic history of the past seven days while the current-day model deals with the current day traffic data. Both of the models capture the spatial interrelation from different hop neighborhoods by separately aggregating different hop neighbor representations while temporal dependency is captured via a weighted spatio-temporal aggregation scheme. Again, we added relative positioning to the node's representation so that our model can distinguish traffic pattern variations from the different periods of a day as well as can discriminate different days in a week. The experimental analysis of real-life datasets verifies the effectiveness of our model in capturing the periodicity of traffic data. 

\section{Acknowledgements}
This project is supported by ICT Division, Government of Bangladesh, and Independent University, Bangladesh (IUB).

\bibliographystyle{splncs04}
\bibliography{mybibliography}

\end{document}